\documentclass[aps,pre,reprint,amsmath,amssymb,nofootinbib]{revtex4-2}
\makeatletter
\def\@hangfrom@section#1#2#3{\@hangfrom{#1#2}#3}
\def\@hangfroms@section#1#2{#1#2}
\makeatother
\usepackage{graphicx}
\usepackage{dcolumn}
\usepackage{bm}
\usepackage{multirow}
\usepackage{booktabs}
\usepackage{xcolor}  
\usepackage{placeins}
\usepackage[colorlinks=true, linkcolor=blue, citecolor=blue, urlcolor=blue]{hyperref}
\begin{document}

\title{Reservoir observer enhanced with residual calibration and attention mechanism}

\author{Yichen Liu}
	\email{yichen\_liu@stu.pku.edu.cn}
\author{Wei Xiao}
\author{Tianguang Chu}
	
\affiliation{School of Advanced Manufacturing and Robotics, Peking University, Beijing 100871, China}

\date{\today}
\begin{abstract}
	\begin{center}
		\begin{minipage}{0.75\linewidth}

	Reservoir observers provide a data-driven approach to the inference of unmeasured variables from observed ones for nonlinear dynamical systems. 
	While previous studies have demonstrated wide applicability, their performance may vary considerably with different input variables, even compromising reliability in the worst cases.
	To enhance the performance of inference, we integrate residual calibration and attention mechanism into the reservoir observer design. 
	The residual calibration module leverages information from the estimation residuals to refine the observer output, and the attention mechanism exploits the temporal dependencies of the data to enrich the representation of reservoir internal dynamics. 
	Experiments on typical chaotic systems demonstrate that our method substantially improves inference accuracy, especially for the worst cases resulting from the traditional reservoir observers. 
	We also invoke the notion of transfer entropy to explain the reason for the input-dependent observation discrepancy and the effectiveness of the proposed method.
	\end{minipage}
\end{center}
\end{abstract}



\maketitle

\section{Introduction}
\begin{figure*}
	\centering  
	\includegraphics[width=\textwidth]{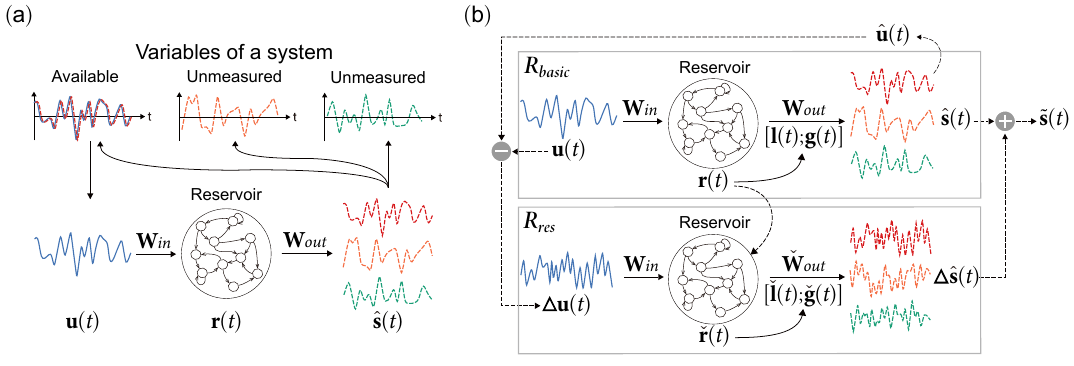}  
	\caption{Schematic diagrams of (a) the traditional reservoir observer and (b) the proposed RORA.}  
	\label{2observer}
\end{figure*}

Observing dynamical systems from partially measured data has consistently been a major focus in many applications.
While various observers based on conventional techniques have proven effective in solving linear system problems \cite{observer}, designing an observer for nonlinear systems remains challenging due to the inherent complexity of nonlinear dynamics.

Recently, reservoir observers (ROs) \cite{ro}, a kind of nonlinear observers based on reservoir computing (RC) \cite{rc}, have emerged as an efficient and model-free approach to estimating unmeasured variables.
Basically, RC is a recurrent neural network training framework that simply requires the weights in the output layer to be trained by regressions, while the weights in the input layer and the reservoir network are generated randomly and then fixed.
The high computational efficiency and simplicity in the RC training process have made it applicable to a wide range of applications, including the prediction of nonlinear time series \cite{prediction1,prediction2,prediction3,prediction4}, the replication of chaotic attractors \cite{attractor1,attractor2}, the classification of nonlinear signals \cite{classification1,classification2,classification3}, chaos synchronization \cite{synchronization1,ro0}, the prediction of critical transitions \cite{tip1,tip2}, the estimation of variables from measured data \cite{ro0,ro1,ro2}, etc. 

In an RO, available observation data are used as input to the reservoir to infer the unmeasured variables, as illustrated in Fig.~\ref{2observer}(a).
The training occurs only at the output layer by minimizing the residual values between the output and the target values.
To date, the RO approach has been employed to estimate unmeasured variables in many studies, e.g., chaos synchronization \cite{ro0}, chaotic fluid systems \cite{ro1}, and spatio-temporal dynamics of excitable media \cite{ro2}.

While significant progress has been achieved so far, certain issues remain to be addressed with the application of the RO approach. 
One is concerned with the possible inconsistent observation performances of an RO for different input variables. 
Of these, the worst cases may not be viewed as reliable due to large inference errors.
For instance, it can be found that using the $z$ variable of the R\"ossler system as input to observe other variables gives poor results, with much lower inference accuracy compared with that obtained by using $x$ or $y$ as input variable. 
Similar cases also occur for other systems, such as the Chua's circuit systems.


To deal with this issue, we focus on the following two aspects. 
First, we intend to exploit the RO's output residuals to improve the observation results.
Notice that in a full state observer, the errors between the input and its estimation in the output are always available in operation. 
This distinctive advantage may be exploited to rectify the observer performance through leveraging additional information in residuals to complement output estimates. 
So far in the literature, little has been done in exploring the potential of residuals explicitly in RO design. 
We will demonstrate that, by carefully incorporating residual information, one can substantially improve the estimation accuracy of ROs.

Another aspect we shall address concerns the temporal dependence of the data processed by ROs. 
In general, traditional ROs exhibit a fading memory, ensuring that the dynamical evolution of the reservoir system remains stable and is independent of its initial conditions \cite{fading}.
On the other hand, however, it may constrain the temporal receptive field of ROs, and thus cause them to focus more on recent states and neglect relatively distant yet temporally informative patterns. 
This would be adverse to obtaining precise results with ROs.
To overcome this, we incorporate the attention mechanism into ROs to take advantage of the inherent temporal correlation of the data.

We note that the attention mechanism was proposed in previous studies for machine translation, allowing neural networks to dynamically attend to the most relevant segments of the input sequence to achieve better semantic alignment \cite{deeplearning}.
Recently, it has been applied to improve the performance of RC. 
For instance, Ref.~\cite{attro1} designed an attention-based leaky function to generate an adaptive RC leakage rate to capture the spatial correlation inherent in figures.
In Ref.~\cite{attro2}, an attention matrix is introduced into the output layer of RC and trained by a gradient-descent algorithm. 
In this work, we propose a simple yet efficient method to leverage the temporal correlation of the data in ROs. 
Specifically, we choose a set of reservoir state points in the past as the centers of attention and compute the attention weights using a Gaussian radial basis function (GRBF). 
For simplicity, the attention centers are selected randomly to retain the merit of the RO framework. 

We demonstrate the effectiveness of the proposed methods with the data generated by the typical R\"ossler, Lorenz, and Chua's circuit systems and spatiotemporal Kuramoto-Sivashinsky equations.  
The results show that a significant improvement in estimation performance can be achieved by our approach, especially for the worst cases where traditional ROs exhibit poor accuracy.
In the presence of measurement noise, the approach appears robust to a certain extent.
Furthermore, we also examine the cause of input-dependent performance discrepancies based on the notion of transfer entropy \cite{te}. 
The analysis suggests that the input variables associated with poor performance usually have weaker influence on the rest variables of the system, thereby limiting the inference capability of ROs.
In such a situation, the proposed attention mechanism can enhance temporal dependency modeling within the reservoir and promote a comprehensive utilization of available input information.

The paper is organized as follows: 
Sec.~\ref{rc} introduces basic reservoir observers.
Section~\ref{method} presents the proposed reservoir observers with residual calibration and attention mechanism (RORA). 
Numerical experiments and discussions are given in Sec.~\ref{exp} and Sec.~\ref{discuss}, respectively.
Section~\ref{conclu} concludes the paper.

\section{Reservoir observers \label{rc}}
The reservoir observer proposed in \cite{ro} employs reservoir computing to infer variables in a nonlinear system with the measurements of available variables as inputs. 
Its structure consists of an input layer, a recurrent high-dimensional reservoir network, called the reservoir layer, and an output layer, illustrated in Fig.~\ref{2observer}(a).
The input and reservoir layers are randomly generated, and the output layer is trainable.
The RO predicts the target variable sequences $\mathbf{s}(t) \in \mathbb{R}^m $ with the input variable sequences $\mathbf{u}(t) \in \mathbb{R}^n$, where $t$ represents discrete time with a constant time interval $\Delta t$.
The reservoir layer comprises $d$ dynamical nodes whose state $\mathbf{r}(t) \in \mathbb{R}^d $ is governed by the rule
\begin{equation}
\mathbf{r}(t+\Delta t) = (1-\alpha)\mathbf{r}(t) +\alpha \tanh\left[\mathbf{Ar}(t) + \mathbf{W}_{\mathrm{in}} \mathbf{u}(t)+\xi \mathbf{1}\right],
\label{rt}
\end{equation}
where $\mathbf{A} \in \mathbb{R}^{d \times d}$ is a sparse weighted adjacency matrix of the reservoir network with the connectivity density $D$, and $\mathbf{W}_{\mathrm{in}} \in \mathbb{R}^{d \times n}$ is the weight matrix of the input layer that couples the input $\mathbf{u}(t)$ to the $d$-dimensional state space. 
The parameter $\alpha \in (0,1]$ is the ``leakage'' rate that controls the speed of the reservoir evolution, and $\xi \mathbf{1}$ is a bias term with $\mathbf{1}=[1,\dots,1]^{\top} \in \mathbb{R}^d$.
Generally, $\mathbf{A}$ is generated randomly with a spectral radius $\rho$, and the elements of $\mathbf{W}_{\mathrm{in}}$ are drawn independently and uniformly from $\left[-\gamma , \gamma\right]$ where $\gamma$ is a scalar.
The prediction of the target $\hat{\mathbf{s}}(t)$ is obtained via a linear transformation
\begin{equation}
\hat{\mathbf{s}}(t) = \mathbf{W}_{\mathrm{out}}\mathbf{r}(t)+\mathbf{b},
\label{st}
\end{equation}
where $\mathbf{W}_{\mathrm{out}} \in \mathbb{R}^{m \times d}$ is the output matrix that maps the reservoir state $\mathbf{r}(t)$ onto the output space, and $\mathbf{b}\in \mathbb{R}^m$ is the output bias.

As a rule, we assume that the target variables can be measured during a limited period $\left[t_1,t_T\right]$ to obtain a well-trained output layer in Eq.~(\ref{st}), by solving the regression problem:
\begin{equation}
\underset{\mathbf{W}_{\mathrm{out}},\mathbf{b}}{\min} \, \left[\sum_{t=t_1}^{t_T}\Vert \mathbf{W}_{\mathrm{out}} \mathbf{r}(t)+ \mathbf{b} - \mathbf{s}(t) \Vert^2 + \beta \text{Tr}(\mathbf{W}_{\mathrm{out}}\mathbf{W}_{\mathrm{out}}^{\top}) \right]	,
\label{min-loss}
\end{equation}
where $\Vert \cdot \Vert$ is the Euclidean norm of vectors and $\beta > 0 $ is the regularization hyperparameter to penalize overfitting.
The optimal solution is calculated by
\begin{equation}
\mathbf{W}_{\mathrm{out}}^* = \mathbf{SR}^{\top}(\mathbf{RR}^{\top}+ \beta \mathbf{I})^{-1},
\label{solu}
\end{equation}
\begin{equation}
\mathbf{b}^* = \bar{\mathbf{s}} - \mathbf{W}_{\mathrm{out}}\bar{\mathbf{r}},
\label{solu-b}
\end{equation}
where $\bar{\mathbf{s}}$ and $\bar{\mathbf{r}}$ are the mean of $\mathbf{s}(t)$ and $\mathbf{r}(t)$ respectively for $t \in \left[t_1,t_T\right]$, and $\mathbf{I} \in \mathbb{R}^{d \times d}$ is an identity matrix. 
The target matrix $\mathbf{S}=[\mathbf{s}(t_1)-\bar{\mathbf{s}},\dots,\mathbf{s}(t_T)-\bar{\mathbf{s}}]$ and the state matrix $\mathbf{R}=[\mathbf{r}(t_1)-\bar{\mathbf{r}},\dots,\mathbf{r}(t_T)-\bar{\mathbf{r}}]$.
 
\section{Enhanced reservoir observers\label{method}}
This section presents our approach for enhancing the performance of reservoir observers through residual calibration and attention mechanism. 

\begin{figure}
	\centering  
	\includegraphics[width=\columnwidth]{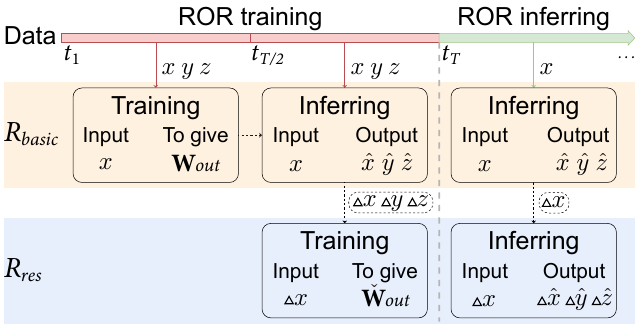}  
	\caption{Schematic diagram of the residual calibration approach.}  
	\label{ror}
\end{figure}

\subsection{Residual calibration\label{res}}
In reservoir observers the residuals of the available variables can be obtained along with the estimation of the unmeasured variables.\footnote{To be more exact, the residuals may come from the regression errors and the measurement noise in raw data.}
This differs from the case of usual prediction problems and may be exploited to improve the reservoir observer performance. 
Here we propose a residual calibration approach to extract additional information from the residuals of the observables to complement the output estimates. 
The idea is to integrate a residual RO module with the usual RO to process the residuals between the input and its estimation in the RO outputs, and thereby to rectify the observation results, as shown schematically in Fig.~\ref{2observer}(b).

To be specific, the basic RO module $R_{\mathrm{basic}}$ takes the traditional form described in Sec.~\ref{rc} with the measured variables as the input $\mathbf{u}(t)$, and the output $\hat{\mathbf{s}}(t)$ gives full state observation of all variables, which contains the estimation of $\mathbf{u}(t)$, denoted by $\hat{\mathbf{u}}(t)$.
Let $\Delta \mathbf{u}(t) = \mathbf{u}(t) - \hat{\mathbf{u}}(t)\in \mathbb{R}^n$ be the residuals of $\mathbf{u}(t)$.
The residual RO module $R_{\mathrm{res}}$ operates according to the following equation
\begin{equation}
	\begin{aligned}
	\check{\mathbf{r}}(t+\Delta t) = (1-&\alpha)\check{\mathbf{r}}(t)  +\alpha \tanh[ \lambda\mathbf{B}\check{\mathbf{r}}(t) \\
	&+ (1-\lambda)\mathbf{Ar}(t) + \mathbf{W}_{\mathrm{in}} \Delta \mathbf{u}(t)+\xi \mathbf{1}],
	\label{rrt}
	\end{aligned}
\end{equation}
where $\check{\mathbf{r}}(t) \in \mathbb{R}^d$ and $\mathbf{B} \in \mathbb{R}^{d \times d}$ are the reservoir state and the adjacency matrix of the reservoir network in $R_{\mathrm{res}}$.
The parameter $\lambda \in (0,1]$ specifies the coupling strength between the two modules, which is used to associate the processing of residual data with that of the original data and ensure the data processed in the two modules relevant in the whole system.
The coupling matrix $\mathbf{A}$ is the same as in Eq.~(\ref{rt}).
Our experiment results show that such a choice of the matrix $\mathbf{A}$ does not significantly affect the system performance and helps in preserving a certain degree of dynamical consistency between the two modules.
All remaining terms are identical to those in Eq.~(\ref{rt}). 
The output of $R_{\mathrm{res}}$ reads 
\begin{equation}
	\Delta\hat{\mathbf{s}}(t) = \check{\mathbf{W}}_{\mathrm{out}}\check{\mathbf{r}}(t)+\check{\mathbf{b}},
	\label{dst}
\end{equation}
where $\check{\mathbf{W}}_{\mathrm{out}}\in \mathbb{R}^{m \times d}$ and $\check{\mathbf{b}}\in \mathbb{R}^m$ are the output matrix and the bias of $R_{\mathrm{res}}$ respectively, which are trained in the same way as the case of $R_{\mathrm{basic}}$.
The final observation results are given by $\tilde{\mathbf{s}}(t) = \hat{\mathbf{s}}(t)+\Delta\hat{\mathbf{s}}(t)$. 
We refer to the above method as reservoir observer with residual calibration (ROR) hereafter.

In application of ROR, $R_{\mathrm{basic}}$ and $R_{\mathrm{res}}$ are trained successively on different segments of the data measured on the interval $\left[t_1,t_T\right]$. 
Specifically, $R_{\mathrm{basic}}$ is trained first on the interval $\left[t_1,t_{T/2}\right]$ and is then applied for inference on the interval $\left[t_{T/2},t_T\right]$ to obtain the errors between the measured values and the estimations, which are finally used to train $R_{\mathrm{res}}$ on the interval $\left[t_{T/2},t_T\right]$.
We illustrate this procedure in Fig.~\ref{ror} for a dynamical system with three variables $x$, $y$, and $z$ [i.e., $\mathbf{s}(t)$], where $x$ is the measured variable [i.e., $\mathbf{u}(t)$].

The training process of $R_{\mathrm{res}}$ is similar to that of $R_{\mathrm{basic}}$, with the columns of matrices $\mathbf{S}$ and $\mathbf{R}$ in Eq.~(\ref{solu}) and the means $\bar{\mathbf{s}}$ and $\bar{\mathbf{r}}$ in Eq.~(\ref{solu-b}) replaced by the corresponding forms of $\Delta\mathbf{s}(t)$ and  $\check{\mathbf{r}}(t)$, respectively.
In addition, $R_{\mathrm{res}}$ employs the same regularization hyperparameter $\beta$ as $R_{\mathrm{basic}}$ in Eq.~(\ref{solu}). A detailed explanation of this choice is given in Appendix~\ref{beta}.

Before ending this section, we remark that another possible training scheme for ROR is to train both $R_{\mathrm{basic}}$ and $R_{\mathrm{res}}$ over the full interval $[t_{1},t_{T}]$, but this strategy degrades the generalization ability of $R_{\mathrm{res}}$, leading to inferior overall performance. A more detailed discussion about this is provided in Appendix~\ref{rp}.

\subsection{Attention mechanism\label{attention}}

Usually, the output of RO at a time instant is a linear projection of the reservoir state at the same time, regarding little about the temporal dependence of the data sequence explicitly. 
To explore the potential of the temporal correlation inherent in the data, we incorporate the attention mechanism into the RO reservoir layer to allow for referencing previous reservoir states in the estimation of the variables.

Here we employ the GRBF to evaluate the attention weights. In this case, the computational cost of the attention on all historical states will increase linearly with the sequence length \cite{deeplearning}. 
To reduce the computational burden, we instead calculate sparse attention weights by limiting the attention range to a set of $N_c$ selected historical states $\mathbf{c}_1,\dots,\mathbf{c}_{N_c}$.
The attention weights of a reservoir state $\mathbf{r}(t)$ to the selected states are defined as:
\begin{equation}
	\phi_{i}(t) = \text{exp}\left[ -\frac{\Vert\mathbf{r}(t) - \mathbf{c}_i \Vert ^2}{2\sigma^2} \right], 
	\label{att}
\end{equation}
where $\mathbf{c}_i$ denotes the $i$th center of the GRBF and $\sigma$ is the bandwidth of the function, representing the influence range of $\mathbf{c}_i$.
Intuitively, the weight $\phi_{i}(t)$ may be viewed as a nonlinear metric for the distances between the state $\mathbf{r}(t)$ and the historical reference states $\mathbf{c}_i$ in reservoir state space. 
The attention vector of $\mathbf{r}(t)$ is then given by 
\begin{equation}
	\mathbf{g}(t) = \frac{1}{N_c}\sum_{i=1}^{N_c} \phi_i(t) \mathbf{c}_i,
	\label{gt}
\end{equation}
where $\mathbf{g}(t) \in \mathbb{R}^d$.
There are several ways to choose the centers, such as random selection, clustering algorithms, and optimization algorithms \cite{deeplearning,centers}. 
Here, we adopt the random selection method due to its simplicity and efficiency.

\begin{figure}[b]
	\centering  
	\includegraphics[width=\columnwidth]{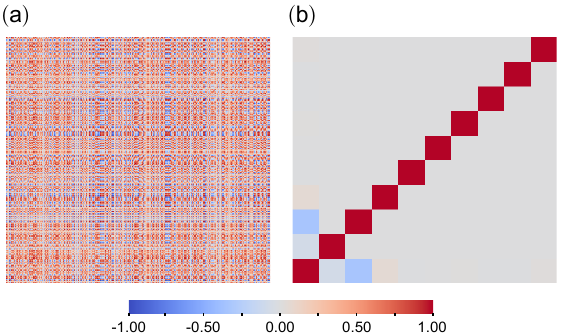}  
	\caption{Correlation matrices of the elements of (a) the reservoir state and (b) the corresponding principal components obtained via SVD for the input variable $x$ in the R\"ossler system.}  
	\label{r corr}
\end{figure}

\begin{table*}
	\caption{\label{hyperparameter}
	Hyperparameters for the systems in experiments.}
	\begin{ruledtabular}
		\begin{tabular}{lcccccccccccc}
System & $T$ & $\Delta t$ & $d$ & $D$ & $\alpha$ & $\xi$ & $\rho$ & $\gamma$ & $\beta$ & $\lambda$ & $\sigma$ & $N_c$  \\ 
\colrule
\addlinespace[2pt]
R\"ossler &  400  &  0.1  &  400  &  0.05  &  1.0  &  1.0  &  1.0  &  1.0  &  $10^{-8}$  &  0.9  &  1.0  &  50   \\
Lorenz &  800  & 0.05  &  400  &  0.05  &  1.0  &  1.0  &  1.0  &  1.0  &  $10^{-8}$  &  0.9  &  1.0  &  50   \\
Chua's circuit &  1000  & 0.1  &  400  &  0.05  &  1.0  &  1.0  &  1.0  &  1.0  &  $10^{-8}$  &0.5  &  1.0  &  50    \\
Kuramoto-Sivashinsky &  30,000  & 0.25  &  1,000  &  0.06  &  0.5  &  0.0  &  0.9  &  0.5  & $10^{-10}$ & 0.95 & 1.0 & 50 
\end{tabular}
\end{ruledtabular}
\end{table*}

We notice that when the number of nodes in the reservoir layer is large enough, the attention weights in Eq.~(\ref{att}) may not well reflect the relevance between the state $\mathbf{r}(t)$ and the center $\mathbf{c}_i$ due to the ``curse of dimensionality'' \cite{curse}.
That is, in a high dimensional embedding space, pairwise distances tend to concentrate and the contrast between relevant and irrelevant states diminishes. 
Therefore, for a reservoir with high state space dimensionality and limited amount of training data, computing the attention vector $\mathbf{g}(t)$ in Eq.~(\ref{gt}) directly in the full state space may give rise to more correlated results and hence impair the attention mechanism.
To deal with this, we resort to the singular value decomposition (SVD) technique to find the principal components of the state sequence and obtain low-dimensional state representations, so as to reduce the dimensionality of reservoir state space in evaluating attention vectors.

First, we make the SVD of state matrix $\mathbf{R}$ in Eq.~(\ref{solu}) as $\mathbf{R} = \mathbf{U\Sigma V}^{\top}$, where the columns of $\mathbf{U}$ and $\mathbf{V}$ are the left and right singular vectors of $\mathbf{R}$ respectively, and $\mathbf{\Sigma}$ is a diagonal matrix with the singular values of $\mathbf{R}$ arranged in descending order on its diagonal.
Define a projection matrix $\mathbf{U}_h\in \mathbb{R}^{d \times h}$ with its columns being the first $h$ columns of $\mathbf{U}$ (corresponding to the first $h$ largest singular values in $\mathbf{\Sigma}$).
Then, for $\mathbf{r}(t) \in \mathbb{R}^d$, $\mathbf{l}(t) = \mathbf{U}_h^{\top}\mathbf{r}(t) \in \mathbb{R}^h$ gives a low-dimensional approximation of the original reservoir state space. 
The value of $h$ can be determined by the optimal-SVHT algorithm \cite{svht}.
With a slight abuse of notations, we use $\mathbf{c}_i \in \mathbb{R}^h$ to signify the selected reference state for attention. 
The corresponding attention vector $\mathbf{g}(t)$ is computed by replacing $\mathbf{r}(t)$ and $\mathbf{c}_i \in \mathbb{R}^d$ in Eqs.~(\ref{att}) and (\ref{gt}) with $\mathbf{l}(t)$ and $\mathbf{c}_i \in \mathbb{R}^h$, respectively.
Now, let us define the attention-augmented state as $\mathbf{p}(t) \triangleq \left[ \mathbf{l}(t); \mathbf{g}(t)\right] \in \mathbb{R}^{2h}$, a concatenation of $\mathbf{l}(t)$ and $\mathbf{g}(t)$.
The estimation of $\mathbf{s}(t)$ in Eq.~(\ref{st}) is modified as: 
\begin{equation}
\hat{\mathbf{s}}(t) = \mathbf{W}_{\mathrm{out}} \mathbf{p}(t) + \mathbf{b},
\label{st2}
\end{equation}
where $\mathbf{W}_{\mathrm{out}} \in \mathbb{R}^{m \times 2h}$.
The corresponding optimal solution is given by:
\begin{equation}
\mathbf{W}_{\mathrm{out}}^* = \mathbf{SP}^{\top}(\mathbf{PP}^{\top}+ \beta \mathbf{I})^{-1},
\label{solu2}
\end{equation}
where $\mathbf{S}$ is the target matrix and $\mathbf{P} \in \mathbb{R}^{2h\times T}$ is the attention-augmented state matrix with $\mathbf{p}(t)-\bar{\mathbf{p}}$ as its columns [$\bar{\mathbf{p}}$ is the mean of $\mathbf{p}(t)$ for $t \in \left[t_1,t_T\right]$].
The solution to $\mathbf{b}^*$ is similar to that in Eq.~(\ref{solu-b}), with $\bar{\mathbf{r}}$ replaced by $\bar{\mathbf{p}}$.

As an illustration, we consider the RO with 400 nodes, using the variable $x$ of the R\"ossler system as its input. 
Figure~\ref{r corr}(a) shows the correlation matrix of the elements of the reservoir state corresponding to the data over the interval $\left[t_1,t_{400}\right]$. 
The large red and blue blocks indicate strong correlations among the elements of the reservoir state vector, which cause the pairwise distances between state vectors to collapse into a narrow range. 
In this case, deriving the meaningful attention vector directly based on these vectors may be difficult, since the concentrated distances make it hard to distinguish each other well. 
Fig.~\ref{r corr}(b) gives the correlation matrix corresponding to the principal components of the reservoir state obtained through SVD with $h=10$. 
It shows that the SVD technique can eliminate well the state correlations and yield more distinguishable distances of the states for implementation of the attention mechanism. 

In the RO framework, the attention vector $\mathbf{g}(t)$ contains certain historical state information and thus helps in preserving long-range temporal dependencies of the data that may fade in the current state. 
As a result, the attention-augmented state $\mathbf{p}(t)$ in Eq.~(\ref{st2}) can be more informative for the regression of the observer.
We refer to the attention-equipped reservoir observer as ROA. 
For the general case of RORA, both $R_{\mathrm{basic}}$ and $R_{\mathrm{res}}$ incorporate the attention mechanism, as shown in Fig.~\ref{2observer}(b), and select the respective centers from their training time span of length $T/2$.

We remark that for data contaminated by noise, long-term past measurements would be of gradually diminishing values in inferring unmeasured variables because of noise-induced error accumulation as well as their reducing temporal dependency on the current data.
In this case, one may consider a ``shift-window''-like method by regularly updating the attention centers to mitigate the adverse effect of outdated data.


\section{Experiments\label{exp}}
In this section we conduct experiments on three typical chaotic systems and a spatiotemporally chaotic system, comparing the performance of the methods, including the traditional RO \cite{ro}, ROR, ROA, and RORA introduced in Sec.~\ref{method}.  
Particularly, the methods are assessed on the same dataset in terms of the mean square error (MSE), and all results are averaged over 100 runs, with each run employing a newly initialized set of random input and reservoir layers. 
The hyperparameters for each system are given in Table~\ref{hyperparameter}. 
For convenience in statement, we account for the study with the method RORA in the sequel; the other cases are similar. 

All results in this section are reported relative to the RO baseline.  
Also, since ROR introduces an extra reservoir and ROA appends an attention vector to the state in the reservoir layer, we provide additional experiments in Appendix~\ref{baseline} for comparison.

\begin{table*}
	\caption{\label{rossler results}%
	MSE values and percentage reductions compared to RO for different methods on the R\"ossler system. Results are grouped by different input variables and the corresponding inferred outputs.}
	\begin{ruledtabular}
		\begin{tabular}{lcccccc}
			\multirow{2}{*}{\textrm{Method}} & \multicolumn{2}{c}{\textrm{Input $x(t)$} } & \multicolumn{2}{c}{\textrm{Input $y(t)$}} & \multicolumn{2}{c}{\textrm{Input $z(t)$}}\\ \cline{2-3} \cline{4-5} \cline{6-7} 
			&$y$&$z$&$x$&$z$&$x$&$y$\\ 
			\colrule
			\addlinespace[2pt]
			\multirow{2}{*}{\textrm{RO}}  & $8.12\times10^{-4}$ & $7.51\times10^{-4}$ & $5.74\times10^{-4}$ & $5.99\times10^{-2}$ & $14.56$ & $76.44$ \\
                              &  &  &  &  &  &  \\

\multirow{2}{*}{\textrm{ROR}} & $2.71\times10^{-4}$ & $1.94\times10^{-4}$ & $2.85\times10^{-4}$ & $2.99\times10^{-2}$ & $6.33$ & $27.68$ \\
                              & ($66.63\%$) & ($74.17\%$) & ($50.35\%$) & ($50.08\%$) & ($56.52\%$) & ($63.79\%$) \\

\multirow{2}{*}{\textrm{ROA}} & $2.88\times10^{-4}$ & $1.79\times10^{-4}$ & $2.71\times10^{-4}$ & $1.84\times10^{-2}$ & $0.67$ & $8.28$ \\
                              & ($64.53\%$) & ($76.17\%$) & ($52.79\%$) & ($69.28\%$) & ($95.40\%$) & ($89.17\%$) \\

\multirow{2}{*}{\textrm{RORA}} & $1.50\times10^{-4}$ & $9.05\times10^{-5}$ & $1.56\times10^{-4}$ & $1.66\times10^{-2}$ & $0.13$ & $2.54$ \\
                               & ($81.53\%$) & ($87.95\%$) & ($72.82\%$) & ($72.29\%$) & ($99.11\%$) & ($96.68\%$) \\

		\end{tabular}
	\end{ruledtabular}
\end{table*}

\subsection{R\"ossler system\label{ross}}
We first consider the chaotic R\"ossler system \cite{ottchaos}:
\begin{equation}
\begin{aligned}
\dot{x} &= -y-z, \\
\dot{y} &= x+ay, \\
\dot{z} &= b+z(x-c),
\label{rossler}
\end{aligned}
\end{equation}
where $a=0.5$, $b=2.0$, and $c = 4.0$.
The measurements of one variable are assumed available as the input $\mathbf{u}(t)$ for reservoir computing, and the target $\mathbf{s}(t)$ is the full state of all variables.
The input sequence, taking $\mathbf{u}(t)=x(t)$ for instance [similar for $y(t)$ and $z(t)$], is normalized as
\begin{equation}
	x(t)^*= \frac{x(t) - \langle x(t) \rangle}{\left\langle \left[ x(t) - \langle x(t) \rangle \right]^2 \right\rangle^{1/2}},
	\label{normalize}
\end{equation}
where $\langle\cdot\rangle$ is the time average.
We train RO and ROA using data over $\left[t_1,t_{T}\right]$, while for ROR and RORA, $R_{\mathrm{basic}}$ and $R_{\mathrm{res}}$ are trained on $\mathbf{s}(t)$ within $\left[t_1,t_{T/2}\right]$ and $\Delta\mathbf{s}(t)$ within $\left[t_{T/2},t_{T}\right]$, respectively.
The inference begins after $t_{T}$.
The input setting, normalization, and training process are the same for other systems below.

To make a quantitative comparison for the performance of different methods, Table~\ref{rossler results} presents the average MSE and the corresponding percentage reductions for different types of ROs with every available input variable. 
It is worth noticing that, in the worst case when using $z$ as the input variable, the MSE values of RO outputs are remarkably larger than those of the case for $x$ or $y$ as the input variable and may not be regarded as a reliable observation given the largest value of MSE exceeding 70. 
For this case, both the residual calibration and the attention mechanism can improve the observers' outcomes greatly, and combining the two methods enables RORA to achieve the smallest MSE values of 0.13 and 2.54 for $x$ and $y$ respectively, reducing the error by up to 99.11\% as opposed to that of RO. 

\begin{table*}
	\caption{\label{lorenz results}%
		MSE values and percentage reductions compared to RO for different methods on the Lorenz system. Results are grouped by different input variables and the corresponding inferred outputs.}
		\begin{ruledtabular}
			\begin{tabular}{lcccccc}
				\multirow{2}{*}{\textrm{Method}} & \multicolumn{2}{c}{\textrm{Input $x(t)$} } & \multicolumn{2}{c}{\textrm{Input $y(t)$}}& \multicolumn{2}{c}{\textrm{Input $z(t)$}} \\ \cline{2-3} \cline{4-5}  \cline{6-7}  
				\addlinespace[2pt]
				&$y$&$z$&$x$&$z$&$x^2$ &  $y^2$\\  
				\colrule
				\addlinespace[2pt]
				\multirow{2}{*}{\textrm{RO}}  
	& $5.04\times10^{-3}$ & $3.01\times10^{-2}$ & $1.52\times10^{-6}$ 
	& $4.11\times10^{-2}$ & $9.82\times10^{-7}$ & $1.35\times10^{-4}$ \\
	&  &  &  &  &  &  \\
	
	\multirow{2}{*}{\textrm{ROR}}  
	& $2.45\times10^{-3}$ & $1.31\times10^{-2}$ & $4.68\times10^{-7}$ 
	& $1.24\times10^{-2}$ & $3.77\times10^{-7}$ & $6.99\times10^{-5}$ \\
	& ($51.39\%$) & ($56.48\%$) & ($69.21\%$) & ($69.83\%$) & ($61.61\%$) & ($48.22\%$) \\
	
	\multirow{2}{*}{\textrm{ROA}}  
	& $8.11\times10^{-4}$ & $9.64\times10^{-3}$ & $7.57\times10^{-7}$ 
	& $1.05\times10^{-2}$ & $5.59\times10^{-8}$ & $6.29\times10^{-6}$ \\
	& ($83.91\%$) & ($67.97\%$) & ($50.20\%$) & ($74.45\%$) & ($94.31\%$) & ($95.34\%$) \\
	
	\multirow{2}{*}{\textrm{RORA}}  
	&$1.62\times10^{-4}$ &$6.74\times10^{-3}$& $3.21\times10^{-7}$ &$9.97\times10^{-3}$  &$3.43\times10^{-8}$ &$2.90\times10^{-6}$ \\ 
	& ($96.79\%$) & ($77.61\%$) & ($78.88\%$) & ($75.74\%$) & ($96.51\%$) & ($97.85\%$) \\
	
			\end{tabular}
		\end{ruledtabular}
	\end{table*}

\begin{figure}[b]
	\centering  
	\includegraphics[width=\columnwidth]{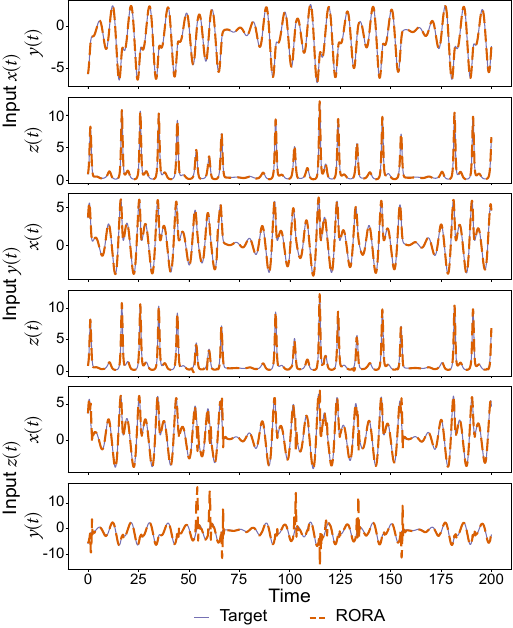}  
	\caption{Inference of variables in the R\"ossler system using RORA.}  
	\label{rossler result}
\end{figure}
\begin{figure}[b]
	\centering  
	\includegraphics[width=\columnwidth]{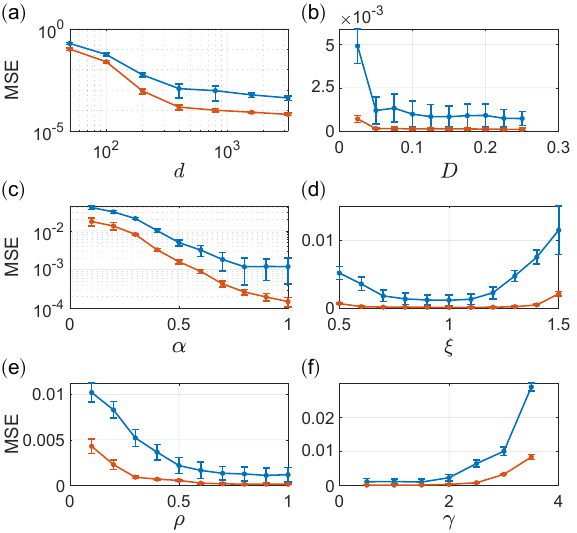}  
	\caption{MSE values of RO (blue) and RORA (red) versus the reservoir hyperparameters. In each case, one hyperparameter is varied while all others are held fixed as given in the previous experiments. }  
	\label{parameter}
\end{figure}
Also, even in the case of using $x$ or $y$ as the input variable, where RO gives relatively accurate estimations, the proposed methods can still effectively reduce the estimation errors. 
Figure~\ref{rossler result} illustrates the inference results of RORA in comparison with the target values, which agree with the results in Table~\ref{rossler results}.
It can be seen from Fig.~\ref{rossler result} that unlike the case of using $x$ or $y$ as the input variable, there exist abrupt deviations from the target values in the RORA estimation curves for the input variable $z$.  
We will discuss on this further in Sec.~\ref{discuss}.

To evaluate the sensitivity of RORA to hyperparameters, we conduct experiments on the R\"ossler system and report in Fig.~\ref{parameter} the corresponding MSE values of both RO and RORA with respect to six major hyperparameters, where $x$ is used as the input to infer $y$. 
For the hyperparameters, RORA consistently outperforms RO over the entire tested range, demonstrating its robustness to the changes of hyperparameter settings. 
In particular, for the hyperparameters $\alpha$ and $\rho$ that dominate the memory capacity and dynamical stability of the reservoir layer, there exist regions, $[0.8, 1.0]$ for $\alpha$ and $[0.6, 1.0]$ for $\rho$, within which RORA maintains stable performance with low error levels.
Appendix~\ref{beta} also gives a discussion concerning the influence of the regularization parameter $\beta$ on the residuals produced by $R_{\mathrm{basic}}$ in ROR.

\subsection{Lorenz system \label{lor}}

\begin{table*}
	\caption{\label{scroll results}%
	MSE values and percentage reductions compared to RO for different methods on the Chua's circuit system. Results are grouped by different input variables and the corresponding inferred outputs.}
	\begin{ruledtabular}
		\begin{tabular}{lcccccc}
			\multirow{2}{*}{\textrm{Method}} & \multicolumn{2}{c}{\textrm{Input $x(t)$} } & \multicolumn{2}{c}{\textrm{Input $y(t)$}} & \multicolumn{2}{c}{\textrm{Input $z(t)$}}\\ \cline{2-3} \cline{4-5} \cline{6-7} 
			&$y$&$z$&$x$&$z$&$x$&$y$\\ 
			\colrule
			\addlinespace[2pt]
			\multirow{2}{*}{\textrm{RO}}  
& $6.25\times10^{-5}$ & $2.14\times10^{-3}$ & 68.31 & 70.82 & $6.72\times10^{-3}$ & $2.47\times10^{-5}$ \\
&  &  &  &  &  &  \\

\multirow{2}{*}{\textrm{ROR}}  
& $9.40\times10^{-6}$ & $2.51\times10^{-4}$ & 21.25& 23.60 & $2.25\times10^{-3}$ & $8.21\times10^{-6}$ \\
& ($84.96\%$) & ($88.27\%$) & ($68.89\%$) & ($66.68\%$) & ($66.52\%$) & ($66.76\%$) \\

\multirow{2}{*}{\textrm{ROA}}  
& $2.25\times10^{-5}$ & $5.42\times10^{-4}$ & 9.59& 10.66& $1.39\times10^{-3}$ & $8.11\times10^{-6}$ \\
& ($64.00\%$) & ($74.67\%$) & ($85.96\%$) & ($84.95\%$) & ($79.32\%$) & ($67.17\%$) \\

\multirow{2}{*}{\textrm{RORA}}  
& $2.16\times10^{-6}$ & $1.31\times10^{-4}$ & 2.17& 2.18& $1.10\times10^{-3}$ & $4.82\times10^{-6}$ \\
& ($96.54\%$) & ($93.88\%$) & ($96.82\%$) & ($96.92\%$) & ($83.63\%$) & ($80.49\%$) \\

		\end{tabular}
	\end{ruledtabular}
\end{table*}

We now investigate the case of the Lorenz system \cite{ottchaos}:
\begin{equation}
\begin{aligned}
\dot{x} &= a(y - x), \\
\dot{y} &= x(b - z) -y, \\
\dot{z} &= cz + xy ,
\label{lorenz}
\end{aligned}
\end{equation}
where $a = 10$, $b = 28$, and $c=-8/3$.
We proceed as in Sec.~\ref{ross} to train the observers for each method.

\begin{figure}[b]
	\centering  
	\includegraphics[width=\columnwidth]{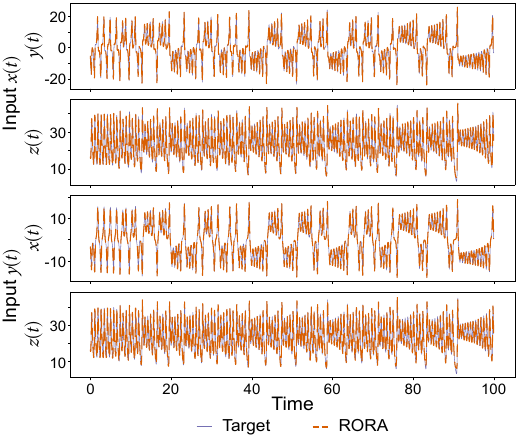}  %
	\caption{Inference of variables in the Lorenz system using RORA.}  
	\label{lorenz result}
\end{figure}

A special property concerning the observation problem of the Lorenz system is its symmetry under the coordinate change of $(x,y,z)$ to $(-x,-y,z)$, which renders the system essentially ``unobservable'' with the measurement of variable $z$. 
In view of this, one may consider inferring the values of $x^2$ and $y^2$, instead of $x$ and $y$ with $z$ as the input variable \cite{ro}. 
For this case, the results in Table~\ref{lorenz results} show that, compared with RO, the proposed RORA can reduce the MSE values of estimations by up to two orders of magnitude. 
For the case of inputting with variable $x$ or $y$, the values of MSE in Table~\ref{lorenz results} confirm the general outperformance of the proposed ROR, ROA, and RORA in comparison with RO. 
The best results are obtained by RORA, with the average MSE of inference an order of magnitude lower than that by RO. 
Especially, when taking $x$ as input variable, RORA realizes a substantial drop in MSE by up to 96.79\% for estimation of variable $y$, compared with RO.
Figure~\ref{lorenz result} plots the observation values inferred from the input variables $x$ and $y$ against their target values. 
It shows clearly that a well-trained RORA can perform the estimation very precisely. 

\begin{figure}[b]
	\centering  
	\includegraphics[width=\columnwidth]{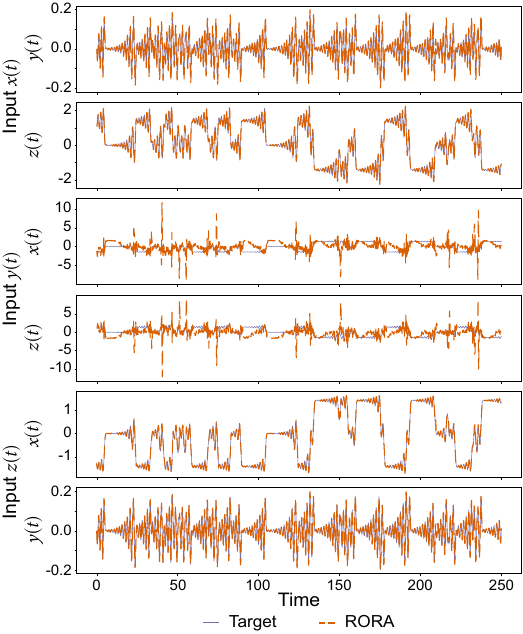}  %
	\caption{Inference of variables in the Chua's circuit system using RORA.}  
	\label{scroll result}
\end{figure}

\subsection{Chua's circuit system\label{scr}}
Our third example concerns the three-scroll Chua's circuit system \cite{scroll3}:
\begin{equation}
\begin{aligned}
\dot{x} &= a[y-g(x)], \\
\dot{y} &= x-y+z, \\
\dot{z} &= by,
\label{scroll}
\end{aligned}
\end{equation}
where $g(x) = c_1 x+c_2 x\vert x \vert + c_3 x^3$, $a=12.8$, $b=-19.1$, $c_1 = 0.6$, $c_2 = -1.1$, and $c_3 = 0.45$.
We carry out the experiment similarly to before.

\begin{table*}
	\caption{\label{ks results}%
	MSE values and percentage reductions ($\downarrow$) compared to RO for different methods on the Kuramoto-Sivashinsky equation. In the setting $\mathrm{IS}_q$, the data used in each of 100 runs are randomly selected from the datasets corresponding to $q$ initial conditions.}
	\begin{ruledtabular}
	\begin{tabular}{clcccccccc}
		\multirow{2}{*}{\begin{tabular}[c]{@{}c@{}}Initialization\\ setting \end{tabular}}&\multirow{2}{*}{\textrm{Method}} & \multicolumn{2}{c}{$n=2$} & \multicolumn{2}{c}{$n=4$} & \multicolumn{2}{c}{$n=6$}& \multicolumn{2}{c}{$n=8$}\\\cline{3-4} \cline{5-6} \cline{7-8} \cline{9-10}
		
		\addlinespace[2pt]
	 & & MSE & $\downarrow$ & MSE & $\downarrow$ & MSE & $\downarrow$ & MSE &$\downarrow$  \\ 
	\colrule
	\addlinespace[2pt]
	\multirow{4}{*}{$\mathrm{IS}_1$}
	& RO   & 1.0716 &  & 0.5294 &  & 0.1533 &  & 0.0385 &  \\ 
	& ROR  & 0.6521 & $39.15\%$ & 0.3691 & $30.28\%$ & 0.1051 & $31.44\%$ & 0.0129 & $66.49\%$ \\ 
	& ROA  & 0.6377 & $40.50\%$ & 0.2414 & $54.41\%$ & 0.0488 & $68.19\%$ & 0.0059 & $84.68\%$ \\ 
	& RORA & 0.4757 & $55.62\%$ & 0.1449 & $72.64\%$ & 0.0385 & $74.90\%$ & 0.0013 & $96.62\%$ \\ 
	\colrule
	\addlinespace[2pt]
  	\multirow{4}{*}{$\mathrm{IS}_{20}$} 
	& RO   & 1.2025 &  & 0.5682 &  & 0.1767 &  & 0.0439 & \\ 
	& ROR  & 0.7150 & $40.53\%$ & 0.4325 & $23.88\%$ & 0.1141 & $35.43\%$ & 0.0088 & $79.95\%$ \\ 
	& ROA  & 0.7445 & $38.10\%$ & 0.2485 & $56.27\%$ & 0.0654 & $62.99\%$ & 0.0083 & $81.09\%$ \\ 
	& RORA & 0.5618 & $53.28\%$ & 0.1711 & $69.88\%$ & 0.0433 & $75.49\%$ & 0.0025 & $94.31\%$ \\ 
	\end{tabular}
\end{ruledtabular}
\end{table*}

Table~\ref{scroll results} gives values of MSE for different methods, which demonstrate the advantages of ROR, ROA, and RORA over RO in general. 
For the worst case of using $y$ as the input variable, like the case of the R\"ossler system with input variable $z$, RO cannot reliably estimate the unmeasured variables $x$ and $y$ with the MSE values exceeding 60. 
In contrast, the proposed RORA exhibits superior inferring performance, drastically reducing the MSE values to below 3. 
For cases where $x$ and $z$ are used as input variables, RORA also gives the best results, with more than 80.49\% reduction in MSE values compared with RO.  
Figure~\ref{scroll result} illustrates clearly the reliable performance of RORA in estimating unmeasured variables when using $x$ or $z$ as the input $\mathbf{u}(t)$.
For the worst case of taking $\mathbf{u}(t)=y(t)$, the estimated values generally fluctuate around the target values closely, with a few spike-like bursts on the curves. 
This phenomenon will be explained later in Sec.~\ref{discuss}.

\subsection{Kuramoto-Sivashinsky equation}
\begin{figure*}[p]
	\centering  
	\includegraphics[height=\dimexpr\textheight-6\baselineskip]{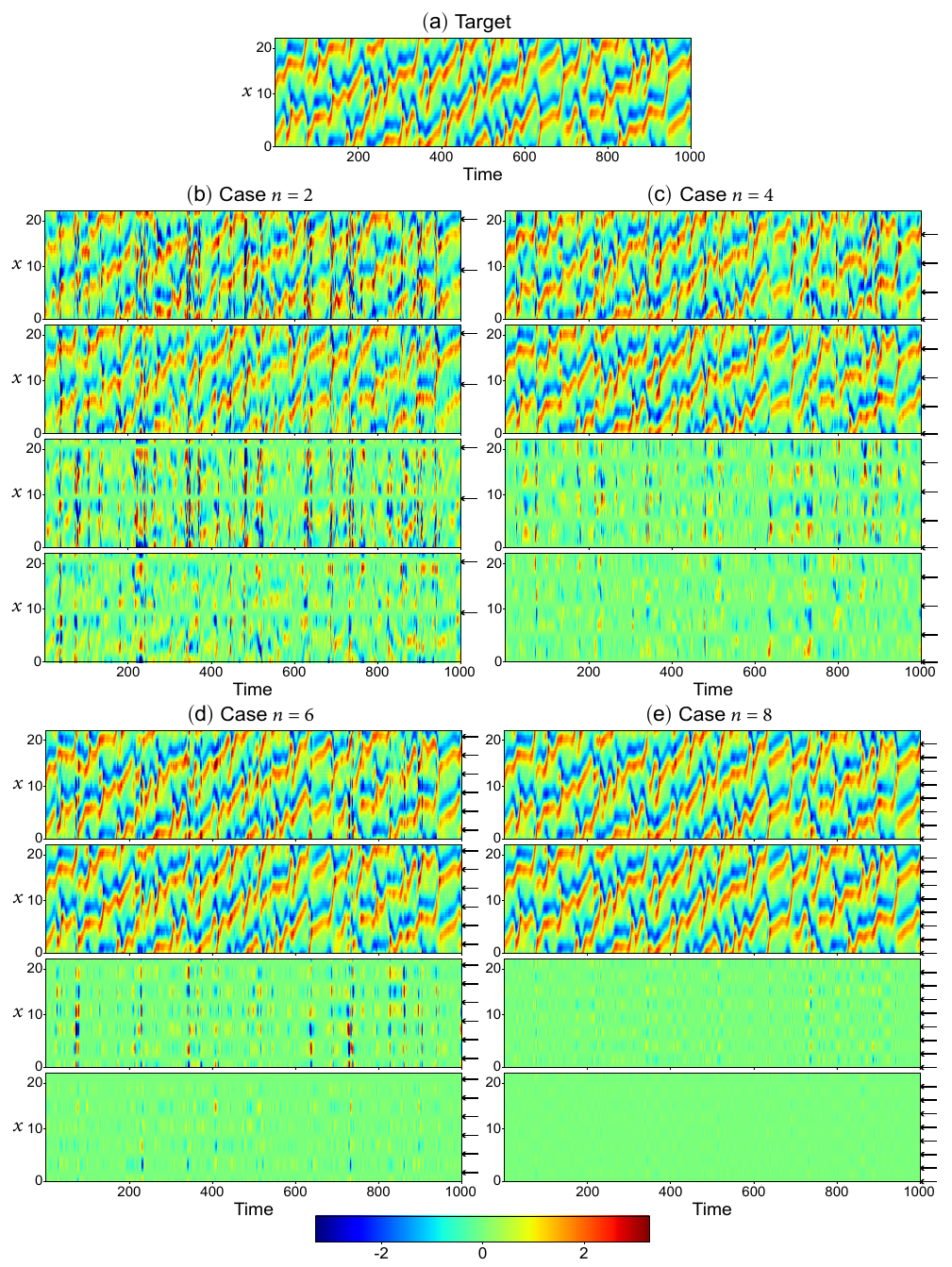}  
	\caption{Inference of the Kuramoto-Sivashinsky equation using RO and RORA. (a) Simulated data from a random initialization. (b--e) Inference results for cases with the number of input variables $n=2, 4, 6,$ and $8$, respectively. In each case, the four panels (from top to bottom) show the RO inference, the RORA inference, and their respective differences from the simulated data. Black arrows to the right of the panels indicate the input locations.}  
	\label{ks result}
\end{figure*}

We further test our methods on spatiotemporally chaotic systems described by the Kuramoto-Sivashinsky equation \cite{ks}
\begin{equation}
	y_t= -yy_x-y_{xx}-y_{xxxx}.
	\label{ks}
\end{equation}
The scalar field $y(x,t)$ is imposed the spatially periodic condition $y(x+L,t)=y(x,t)$ for $x \in [0,L)$, with $L=22$. 
Equation~(\ref{ks}) is integrated on a spatial grid of $Q=64$ equally spaced points, initialized randomly, with a time step of $\Delta t = 0.25$.
The resulting $Q$ sequences serve as the target $\mathbf{s}(t)$.
From this we select $n$ sequences as the input $\mathbf{u}(t)$, where $n= 2,4,6,$ or $8$, setting up four cases, respectively.
The target $\mathbf{s}(t)$ is available during $[t_1,t_T]$ to train the observers.

Fig.~\ref{ks result}(a) shows the simulated data from a random initialization of the Kuramoto-Sivashinsky equation.
The performance of RO and RORA in the four cases is illustrated by Figs.~\ref{ks result}(b)--(e).
In each case, the first two panels depict the inference of RO and RORA, and the last two are their respective differences from the simulated data.
It shows that RORA can infer the simulated data more accurately than RO does in all cases, and the estimation errors decrease very rapidly as $n$ increases. 

Table~\ref{ks results} provides the numerical results for different initialization settings of the Kuramoto-Sivashinsky equation.
In the setting $\mathrm{IS}_q$, the data used in each of 100 runs are randomly selected from the datasets corresponding to $q$ initial conditions. 
For the setting $\mathrm{IS}_1$, the average MSE of RORA is much lower than that of RO in all cases, with both the residual calibration and attention mechanism independently contributing to the improvement in inference accuracy.
In cases with more inputs, the MSE reduction of RORA over RO becomes more pronounced, reaching up to $96.62\%$ when $n=8$.
Furthermore, RORA achieves the same level of accuracy as RO with two additional inputs, alleviating the requirement for extensive measurements when observing all sequences.
For instance, the MSE of RORA with $n=4$ (i.e., 0.1449) is lower than that of RO with $n=6$ (i.e., 0.1533), consistent with the results shown in Fig.~\ref{ks result}.
The results of $\mathrm{IS}_{20}$ are nearly identical to those of $\mathrm{IS}_1$, demonstrating the robust performance of RORA.

\subsection{Effect of measurement noise on residual calibration}

\begin{figure}[t]
	\centering  
	\includegraphics[width=\columnwidth]{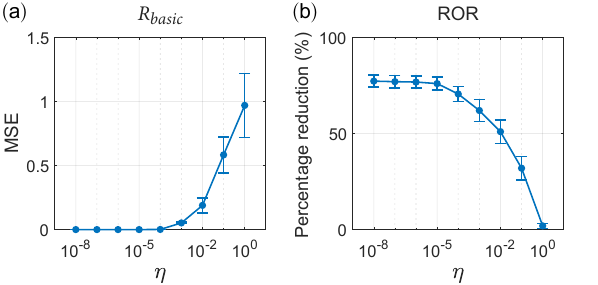}  
	\caption{MSE values of $R_{\mathrm{basic}}$ and the percentage reduction of MSE brought by ROR versus the noise level $\eta$. (a) MSE values against the noise-free sequences under varying levels of measurement noise. (b) Percentage reduction in MSE achieved by ROR compared to RO in the presence of noise.}  
	\label{noise}
\end{figure}

The measurement noise may contaminate the residuals produced by the basic reservoir observer $R_{\mathrm{basic}}$ and consequently affect the performance of ROR.
To investigate this issue, we conduct experiments on inferring $y$ from the input $x$ in the R\"ossler system, where $x$ and $y$ have the ranges $[-3.84,6.32]$ and $[-6.70,2.54]$, respectively.
Specifically, we inject additive noise drawn from a uniform distribution between $-\eta$ and $+\eta$ into the original data sequences.
The noisy measurements are used in both training and inference processes, and the MSE values are evaluated against the corresponding noise-free sequences.
Figure~\ref{noise}(a) demonstrates how the MSE values of $R_{\mathrm{basic}}$, corresponding to the mean of the squared residuals, vary with the noise level.
The results indicate that the residuals remain virtually unaffected for weak noise of certain levels ($\eta<10^{-2}$), and will rise steadily with the increase of noise levels.
Figure~\ref{noise}(b) plots the corresponding percentage reduction in MSE achieved by ROR relative to RO.
Within the range of weak noises, the residual calibration can still reduce the MSE by more than 50\%, and beyond the range the reduction effect gradually diminishes.
When $\eta$ reaches 1, ROR offers no discernible improvement over RO, which indicates that the residual sequences are dominated by noise and no useful information can be extracted.
Similar results can be obtained for the other systems considered in this section.


\section{Discussion\label{discuss}}

\begin{figure*}
	\centering  
	\includegraphics[width=\textwidth]{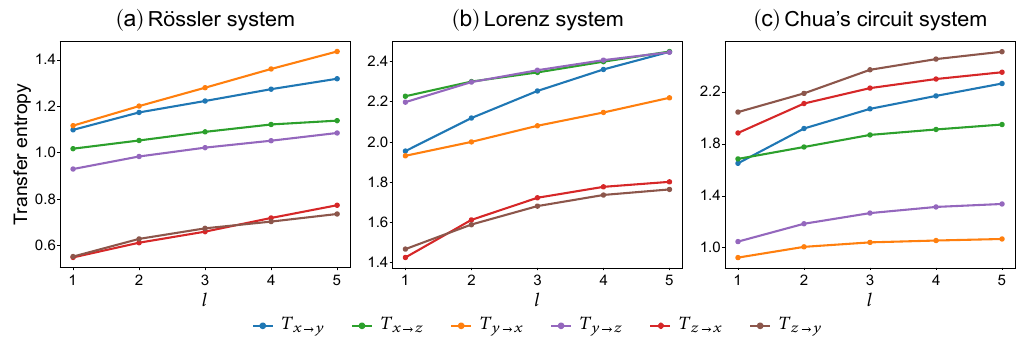}  
	\caption{Transfer entropy between variables in three chaotic systems: (a) the R\"ossler system, (b) the Lorenz system, and (c) the Chua's circuit system. }  
	\label{te3}
\end{figure*}

The above experiment results show that the integration of residual calibration and attention mechanism into the traditional RO can substantially reduce the observation errors in the ``worst cases'' for the chaotic systems under consideration (e.g., inferring variables $x$ and $y$ from input variable $z$ in the R\"ossler system) and hence make the observer more reliable. 
Nevertheless, in these cases, the discrepancies of the observability in terms of MSE values for the enhanced ROR, ROA, and RORA remain obvious, even though the values appear more acceptable than those for RO. 
In order to explore the reason for this issue, here we employ the notion of transfer entropy \cite{te}, which is an information-theoretic measure to quantify the dynamics of the directional information transport from one sequence to the other. 
Unlike symmetric measures such as mutual information, transfer entropy is particularly well suited for serving our purpose because it can quantitatively evaluate the dependency relationship between two variables in a system based on data, hence providing insights into the asymmetry in their interaction. 
In general, the transfer entropy of a sequence $\mathbf{J}$ to the other $\mathbf{I}$ is defined as
\begin{equation}
	T_{\mathbf{J} \to \mathbf{I}} = H\left(\mathbf{I}_{t} \mid \mathbf{I}_{t}^{(k)}\right) - H\left(\mathbf{I}_{t} \mid \mathbf{I}_{t}^{(k)}, \mathbf{J}_{t}^{(l)}\right),
	\label{te}
\end{equation}
where $H( \cdot \mid \cdot)$ represents the conditional entropy, $\mathbf{I}_{t}^{(k)} = (\mathbf{I}_{t-1},\dots,\mathbf{I}_{t-k})$ is the past $k$ states of $\mathbf{I}_{t}$, and $\mathbf{J}_{t}^{(l)} = (\mathbf{J}_{t-1},\dots,\mathbf{J}_{t-l})$ is the past $l$ states of $\mathbf{J}_{t}$.
In \cite{te}, it is typical to take $k=l=1$ in Eq.~(\ref{te}).
By definition, the transfer entropy can be intuitively understood as the degree of dependence of $\mathbf{I}$ on $\mathbf{J}$ by quantifying the additional information provided by $\mathbf{J}$ about $\mathbf{I}$'s future beyond what is already explained by $\mathbf{I}$'s own past. 
A higher value of transfer entropy indicates a stronger information flow from $\mathbf{J}$ to $\mathbf{I}$, i.e., a stronger directional influence of $\mathbf{J}$ on $\mathbf{I}$.

In the context of this work, we can view the input variables and the output targets as $\mathbf{J}$ and $\mathbf{I}$ respectively in Eq.~(\ref{te}). 
According to Eq.~(\ref{rt}), the reservoir state at time $t$ is determined by the input $\mathbf{J}_{t-1}$ and the past state $\mathbf{r}(t-1)$. 
The latter encodes the information from multiple past inputs of $\mathbf{J}_t$ due to the fading memory of the reservoir network dynamics. 
In view of this, here we take $k=1$ and $l=1,\dots,5$ in Eq.~(\ref{te}) to examine the influence of $\mathbf{J}$ on $\mathbf{I}$. 
The results are shown in Fig.~\ref{te3}. 
It is clear that the values of transfer entropy in all cases vary consistently with $l$ and show evident differences in the dependencies among variables in a system. 
In particular, the curves of $T_{z\to x}$ and $T_{z\to y}$ at the bottom are far apart from the other curves in Figs.~\ref{te3}(a) and (b), indicating that variable $z$ contributes minimal information to variables $x$ and $y$ in the R\"ossler and Lorenz systems, and a similar case holds for variable $y$ of the Chua's circuit system as shown in Fig.~\ref{te3}(c). 
Thus, in such cases, the driving variables cannot offer enough correlation information for accurate inference of the unmeasured variables.

On the other hand, however, for the R\"ossler and Chua's circuit systems, despite the limited information afforded by the specific input variable to infer other variables, the attention mechanism can mitigate well the deficiency by incorporating additional reference information. 
Actually, using the GRBF, the attention weights quantify the distances between a reservoir state and its certain past states, namely the attention centers. 
The attention mechanism prioritizes relevant information from the nearby centers and filters out the irrelevant ones, thus assessing dynamical similarity through distance, so as to take into account the most meaningful dynamic patterns for inference. 
This results in a more comprehensive and informative representation of the input, thereby making the inference more reliable and grounded. 

The case for the Lorenz system is quite different from the above mentioned instances. 
The lower informativity of input variable $z$ for inferring other variables, as shown in Fig.~\ref{te3}(b), consists of the inherent reverse invariance for variables $x$ and $y$ in the Lorenz system. 
This cannot be compensated without modifying the structure of the system. 
Alternatively, if one is content with estimating $x^2$ and $y^2$, instead of $x$ and $y$ from the measurement of variable $z$ \cite{ro}, the proposed methods are also capable of yielding more accurate results (see Table~\ref{lorenz results}).

\section{Conclusion\label{conclu}}
Traditional reservoir observers may exhibit remarkable variations in estimation errors with different input variables, including the worst cases that might not be viewed as reliable. 
Considering this deficiency, we introduced the residual calibration module and attention mechanism into the reservoir computing-based nonlinear observer design to enhance the performance of traditional reservoir observers. 
The residual calibration module leverages the estimation residuals to rectify the observation results, and the attention mechanism can exploit temporal dependencies within the data in processing, thus providing more comprehensive information for inference of unmeasured variables. 
Numerical experiments demonstrate that each of the two enhancements can significantly improve the observation results. 
Moreover, their combination gives the best results, with a substantial reduction in mean squared errors up to two orders of magnitude compared with traditional reservoir observers. 
This makes the proposed approach more reliable even in the worst cases, such as the R\"ossler system with input variable $z$ and the Chua's circuit system with input variable $y$, in contrast to previous works.
In the presence of measurement noise, our method can still exhibit certain desirable robust performance.

In addition, we also presented an explanation about the cause of observation inconsistencies for different input variables in terms of transfer entropy. 
According to their transfer entropy values, the underperforming input variables generally have a weaker directional influence on others. 
For such cases, the attention mechanism can incorporate temporal dependencies of the data in reservoir computing to enhance the correlation of the involved variables. 


\appendix

\begin{table*}
	\caption{\label{mean var} Statistics of residuals from $R_{\mathrm{basic}}$ during different time intervals for the R\"ossler system. The intervals $\left[t_{T/2},t_T\right]$ for ROR and $\left[t_1,t_T\right]$ for ROR-al specify the time ranges from which residuals are drawn to train $R_{\mathrm{res}}$, and $\left[t_T,t_{11T}\right]$ denotes the common inference interval for both methods. The mean absolute value (MAV) and variance are calculated on the residuals for each setting ``input $\rightarrow$ target'' and averaged over 100 runs.}
	\begin{ruledtabular}
	\begin{tabular}{llcccccc}
	\multirow{2}{*}{Method}&\multirow{2}{*}{Statistics}&\multicolumn{2}{c}{$x\to y$}& \multicolumn{2}{c}{$y\to z$}&\multicolumn{2}{c}{$z\to x$} \\\cline{3-4} \cline{5-6} \cline{7-8}
	\addlinespace[2pt]
	& & \multicolumn{1}{c}{$\left[t_{T/2},t_T\right]$} & \multicolumn{1}{c}{$\left[t_T,t_{11T}\right]$} & \multicolumn{1}{c}{$\left[t_{T/2},t_T\right]$} & \multicolumn{1}{c}{$\left[t_T,t_{11T}\right]$} & \multicolumn{1}{c}{$\left[t_{T/2},t_T\right]$} & \multicolumn{1}{c}{$\left[t_T,t_{11T}\right]$}  
	\\ \addlinespace[2pt] \hline
	\addlinespace[2pt]
	\multirow{2}{*}{ROR}& MAV & $9.23\times10^{-5}$ & $4.35\times10^{-4}$ &$1.73\times10^{-4}$& $5.63\times10^{-4}$ & $9.70\times10^{-4}$ & $2.78\times10^{-3}$\\
	& Variance  &$4.55\times10^{-8}$ & $2.19\times10^{-7}$ &$ 4.68\times10^{-7}$ &$2.24\times10^{-6}$& $1.03\times10^{-5}$ &$3.85\times10^{-4}$  \\
	\addlinespace[6pt]
	& & \multicolumn{1}{c}{$\left[t_1,t_{T}\right]$}  & \multicolumn{1}{c}{$\left[t_T,t_{11T}\right]$}& \multicolumn{1}{c}{$\left[t_1,t_{T}\right]$}   & \multicolumn{1}{c}{$\left[t_T,t_{11T}\right]$}   & \multicolumn{1}{c}{$\left[t_1,t_{T}\right]$}& \multicolumn{1}{c}{$\left[t_T,t_{11T}\right]$}     \\ \addlinespace[2pt]
	\hline
	\addlinespace[2pt]
	\multirow{2}{*}{ROR-al}&MAV &$6.87\times10^{-5}$   & $3.25\times10^{-4}$ &$8.28\times10^{-5}$  &$5.07\times10^{-4}$ & $7.93\times10^{-4}$  & $2.77\times10^{-3}$\\
	&Variance & $9.71\times10^{-9}$  & $1.67\times10^{-7}$ &$7.35\times10^{-8}$  &$1.45\times10^{-6}$ &$1.75\times10^{-6}$   & $1.26\times10^{-4}$ \\
	\end{tabular}
\end{ruledtabular}
\end{table*}
\begin{table*}
	\begin{ruledtabular}
		\caption{\label{rp result}  
		MSE values of different methods on the Rössler system. Results are grouped by ``input $\rightarrow$ target" pairs, and averaged over 100 runs.}
		\begin{tabular}{lcccccc}
		Method & $x\to y$ & $x\to z$ & $y\to x$ & $y\to z$ & $z\to x$ & $z\to y$\\ \hline
		\addlinespace[2pt]
		ROR    &$2.71\times10^{-4}$&$1.94\times10^{-4}$&$2.85\times10^{-4}$ & $2.99\times10^{-2}$ &6.33 &27.68 \\
		ROR-al &$3.52\times10^{-4}$&$3.26\times10^{-4}$&$3.31\times10^{-4}$ & $3.37\times10^{-2}$ &7.72 &33.68 
		\end{tabular}
	\end{ruledtabular}
	\end{table*}
\begin{figure}[b]
	\centering  
	\includegraphics[width=\columnwidth]{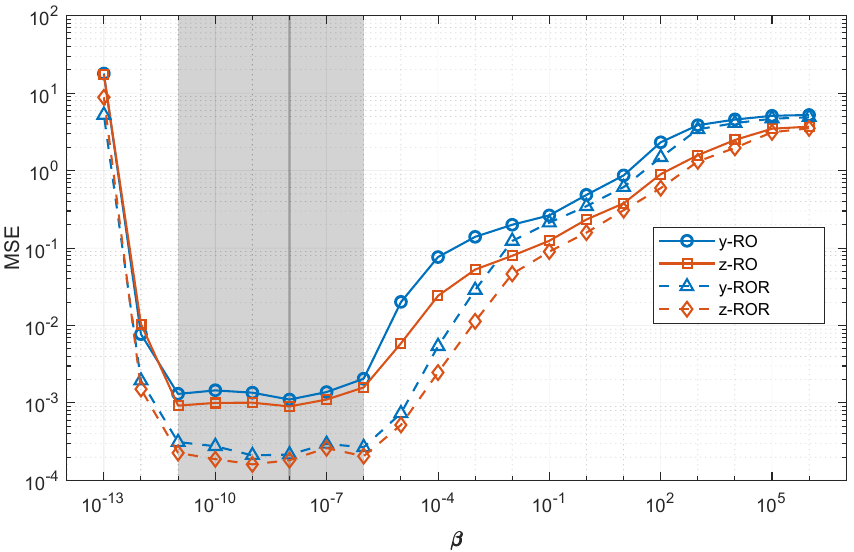}  
	\caption{MSE values of RO and ROR versus the hyperparameter $\beta$ when inferring $y$ and $z$ with inputting $x$ of the R\"ossler system. MSE values are averaged over 100 runs during the inference stage of 2,000 time steps, with the same hyperparameters except $\beta$. For ROR, both reservoirs share the same value of $\beta$. The shaded area indicates the well-regularized region where $\beta$ yields near-optimal performance. Beyond this region, the MSE values of RO and ROR vary over a wide range.}  
	\label{fig beta}
\end{figure}
\section{\texorpdfstring{Effect of $\beta$ on residuals}{Effect of beta on residuals}\label{beta}}
In ROR the regularization hyperparameter $\beta$ in Eq.~(\ref{min-loss}) affects the residuals generated by $R_{\mathrm{basic}}$ when solving for $\mathbf{W}_{\mathrm{out}}$, and hence can have an impact on the performance of the residual reservoir observer $R_{\mathrm{res}}$.
To examine the effect of $\beta$, we plot the MSE curves of RO (equivalent to the mean of the squared residuals from $R_{\mathrm{basic}}$) and ROR over different $\beta$ values in Fig.~\ref{fig beta}.
It shows that there exists a broad well-regularized region ($10^{-11}$ to $10^{-6}$) for $\beta$ where both RO and ROR achieve near-optimal performance. 
Within this region, $R_{\mathrm{basic}}$ remains robust, and $R_{\mathrm{res}}$ appears largely insensitive to variations in $\beta$ values. 
This in turn allows for a better choice of the value for hyperparameter $\beta$ in $R_{\mathrm{res}}$ while maintaining good performance of $R_{\mathrm{basic}}$. 
For convenience we take the same value of $\beta$ in $R_{\mathrm{basic}}$ and $R_{\mathrm{res}}$. 
It is also clear that beyond the well-regularized region, the performance of $R_{\mathrm{basic}}$ and $R_{\mathrm{res}}$ both deteriorates obviously: $\beta \leq 10^{-12}$ leading to overfitting, while $\beta \geq 10^{-5}$ resulting in underfitting.

\begin{figure}[b]
	\centering  
	\includegraphics[width=\columnwidth]{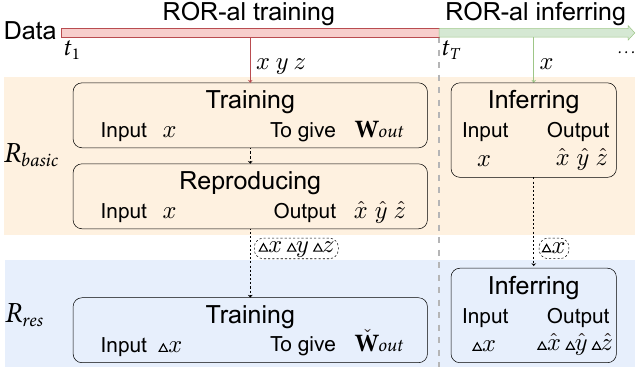}  
	\caption{Schematic diagram of of an alternative residual-calibration scheme ROR-al.}  
	\label{rorrp}
\end{figure}

	\begin{table*}
		\caption{\label{expanding}%
		MSE values of different methods on the three systems. Results are evaluated using the same hyperparameters as in the main text, except for the doubled number of nodes in RO-$2d$, and are grouped by different input variables and the corresponding inferred outputs. For input $z(t)$: $e = 2$ for the Lorenz system and	$e = 1$ otherwise.}
		\begin{ruledtabular}
			\begin{tabular}{llcccccc}
				\multirow{2}{*}{\textrm{Method}}&\multirow{2}{*}{\textrm{System}} & \multicolumn{2}{c}{\textrm{Input $x(t)$} } & \multicolumn{2}{c}{\textrm{Input $y(t)$}} & \multicolumn{2}{c}{\textrm{Input $z(t)$}}\\ \cline{3-4} \cline{5-6} \cline{7-8} 
				\addlinespace[2pt]
				& &$y$&$z$&$x$&$z$&$x^e$&$y^e$\\ 
				\colrule
	
				\addlinespace[2pt]
				\multirow{3}{*}{RO-$2d$}&R\"ossler  &$6.44\times10^{-4}$&$6.65\times10^{-4}$&$4.87\times10^{-4}$ & $5.46\times10^{-2}$ &12.22&60.11 \\
				&Lorenz   &$4.94 \times10^{-3}$&$ 2.88\times10^{-2}$&$9.23\times10^{-7}$ &$4.00\times10^{-2}$ &$8.77\times10^{-7}$&$9.69\times10^{-5}$\\
				&Chua's   &$4.54\times10^{-5}$&$1.18\times10^{-3}$&$61.21$ &$63.87$ &$5.45\times10^{-3}$&$1.31\times10^{-5}$\\
				
	
				\addlinespace[6pt]
				\multirow{3}{*}{P-RC}&R\"ossler   &$6.06\times10^{-4}$&$3.92\times10^{-4}$&$4.79\times10^{-4}$ &$2.57 \times10^{-2}$ &$9.15$&$42.63$\\
				&Lorenz   &$2.73\times10^{-3}$&$2.43\times10^{-2}$&$1.12\times10^{-6}$ &$3.41\times10^{-2}$ &$1.71\times10^{-7}$&$5.79\times10^{-5}$\\
				&Chua's   &$3.62\times10^{-5}$&$1.03\times10^{-3}$&$33.36$ &$40.19$ &$1.73\times10^{-3}$&$9.54\times10^{-6}$\\
				
			\end{tabular}
		\end{ruledtabular}
\end{table*}

\begin{figure*}[!t]
		\centering  
		\includegraphics[width=\textwidth]{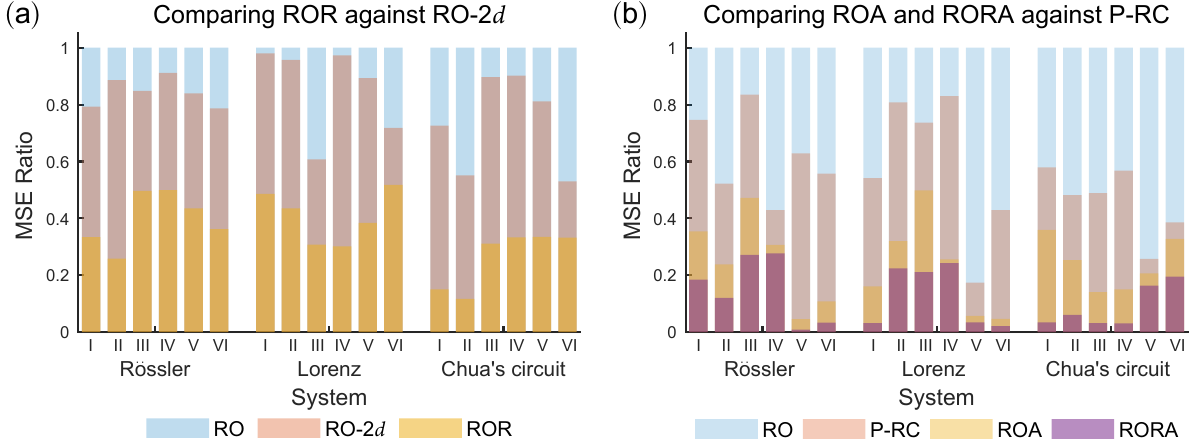}  
		\caption{MSE ratios (normalized by RO) for RO-2$d$, P-RC, and the proposed methods on the concerned systems. (a) Comparison of ROR against RO-2$d$. (b) Comparison of ROA and RORA against P-RC. Roman numerals denote the ``input $\rightarrow$ target" pairs: I: $x\rightarrow y$, II: $x\rightarrow z$, III: $y\rightarrow x$, IV: $y\rightarrow z$, V: $z\rightarrow x$, and VI: $z\rightarrow y$.}  
		\label{baselines}
\end{figure*}

\section{An alternative training scheme of residual calibration\label{rp}}
As described in Sec.~\ref{res}, we train $R_{\mathrm{basic}}$ and $R_{\mathrm{res}}$ in ROR successively on the first and second halves of the data measured on the interval $\left[t_1,t_T\right]$.
Besides this scheme, another possible approach, denoted as ROR-al, is to train both $R_{\mathrm{basic}}$ and $R_{\mathrm{res}}$ over the full interval $[t_1, t_T]$, as illustrated in Fig.~\ref{rorrp}.
Here we provide experimental results to compare the two training schemes to justify our method.

The primary distinction between them lies in the source of the residuals used to train $R_{\mathrm{res}}$. 
Specifically, in our proposed scheme, the residuals are derived from the inference of $R_{\mathrm{basic}}$ over the interval $[t_{T/2},t_T]$, where $R_{\mathrm{basic}}$ estimates the unmeasured variables based on unseen inputs. 
In contrast, the alternative scheme ROR-al computes residuals based on the reproduction of the training data over $[t_1, t_T ]$, where the residuals mostly indicate how well $R_{\mathrm{basic}}$ fits the data it has already seen and limit the potential in developing generalization ability for $R_{\mathrm{res}}$.

Actually, it is recognized in the literature that residuals tied to seen data tend to underestimate future uncertainty and thus provide weaker guidance for improving generalization \cite{generalize}.
Table~\ref{mean var} provides a statistical comparison of residuals obtained by the R\"ossler system with the two schemes. 
The results show that residuals from the inference stage on $[t_{T/2},t_T]$ exhibit larger magnitude and variance, and their distribution more closely resembles those in subsequent intervals; whereas residuals from the training stage on the whole interval $[t_1,t_T]$ are smaller in magnitude, more concentrated in distribution, and less representative of future errors. 
Meanwhile, we compare the performance of ROR and ROR-al on the R\"ossler system in terms of MSE in Table~\ref{rp result}. 
It shows a consistent outperformance of ROR over ROR-al in all cases.



\section{Further results for comparison \label{baseline}}

Notice that compared with traditional RO, the present ROR introduces a residual reservoir module $R_{\mathrm{res}}$ to further process the residuals generated by the basic reservoir $R_{\mathrm{basic}}$. 
This doubles the number of nodes in the entire system.
To compared with the ROR results presented in Tables~\ref{rossler results}--\ref{scroll results}, we conduct several experiments on traditional RO with the same size ($2d$ nodes) and summarize the results in Table~\ref{expanding}. 
It can be seen that the ROR consistently outperforms the RO under the new baseline.

Another comparative study concerns the ROA, which makes use of the reservoir state together with an attention vector for regression during training. 
This introduces a nonlinear effect of the reservoir state in the training process.
Following one anonymous reviewer's suggestion, we examine the case where the state vector $\mathbf{r}(t)$ and its element-wise squared counterpart, denoted as $\mathbf{r}^2(t)$, form a new composite state vector $[\mathbf{r}(t);\mathbf{r}^2 (t)] \in \mathbb{R}^{2d}$ for regression, as in the polynomial reservoir computing (P-RC) approach \cite{expanding}.
Experiment results shown in Table \ref{expanding}, together with those of ROA and RORA in Tables~\ref{rossler results}--\ref{scroll results}, confirm the advantage of both ROA and RORA over the RO with the new setting.

Finally, for an immediate comparison, Fig.~\ref{baselines} plots the MSE ratios normalized by RO, comparing ROR against RO-2$d$ and comparing ROA and RORA against P-RC, on the three typical chaotic systems under consideration.

\bibliography{ref.bib}

\end{document}